\def\year{2020}\relax
\documentclass[letterpaper]{article} % DO NOT CHANGE THIS
\usepackage{aaai20}  % DO NOT CHANGE THIS
\usepackage{times}  % DO NOT CHANGE THIS
\usepackage{helvet} % DO NOT CHANGE THIS
\usepackage{courier}  % DO NOT CHANGE THIS
\usepackage[hyphens]{url}  % DO NOT CHANGE THIS
\usepackage{graphicx} % DO NOT CHANGE THIS
\usepackage{graphicx}  % DO NOT CHANGE THIS
\usepackage{rotating}
\usepackage{tikz}
\usepackage{todonotes}
\usepackage{amssymb}
\usepackage{amsmath}
\usepackage{booktabs}
\usepackage{multirow}
\usepackage{algorithm}
\usepackage{pifont}% http://ctan.org/pkg/pifont
\newcommand{\cmark}{\ding{51}}%
\newcommand{\xmark}{\ding{55}}%
\usepackage[noend]{algpseudocode}
\makeatletter
\let\OldStatex\Statex
\renewcommand{\Statex}[1][3]{%
  \setlength\@tempdima{\algorithmicindent}%
  \OldStatex\hskip\dimexpr#1\@tempdima\relax}
\makeatother
\algnewcommand{\LineComment}[1]{\State \(\triangleright\) #1}
\newcommand*\circled[1]{\tikz[baseline=(char.base)]{
            \node[shape=circle,draw,inner sep=0.55pt] (char) {#1};}}
\title{Knowledge Distillation from Internal Representations }
\author{
    Gustavo Aguilar,\textsuperscript{\rm 1} 
    Yuan Ling,\textsuperscript{\rm 2}
    Yu Zhang,\textsuperscript{\rm 2}
    Benjamin Yao,\textsuperscript{\rm 2}
    Xing Fan,\textsuperscript{\rm 2}
    Chenlei Guo\textsuperscript{\rm 2}\\ 
    \textsuperscript{\rm 1}Department of Computer Science, University of Houston, Houston, USA\\
    \textsuperscript{\rm 2}Alexa AI, Amazon, Seattle, USA\\
    gaguilaralas@uh.edu, \{yualing,yzzhan,banjamy,fanxing,guochenl\}@amazon.com 
}
\begin{document}

\maketitle

\begin{abstract}
Knowledge distillation is typically conducted by training a small model (the student) to mimic a large and cumbersome model (the teacher). The idea is to compress the knowledge from the teacher by using its output probabilities as soft-labels to optimize the student. However, when the teacher is considerably large, there is no guarantee that the internal knowledge of the teacher will be transferred into the student; even if the student closely matches the soft-labels, its internal representations may be considerably different. This internal mismatch can undermine the generalization capabilities originally intended to be transferred from the teacher to the student. In this paper, we propose to distill the internal representations of a large model such as BERT into a simplified version of it. We formulate two ways to distill such representations and various algorithms to conduct the distillation. We experiment with datasets from the GLUE benchmark and consistently show that adding knowledge distillation from internal representations is a more powerful method than only using soft-label distillation.
\end{abstract}

% ========================================================================
\section{Introduction}
% ========================================================================

% PARAGRAPH MESSAGE: complex transformer-based models are really good but expensive
Transformer-based models have significantly advanced the field of natural language processing by establishing new state-of-the-art results in a large variety of tasks. Specifically, BERT \cite{devlin2018bert}, GPT \cite{radford2018improving}, GPT-2 \cite{radford2019language}, XLM \cite{lample2019xlm}, XLNet \cite{yang2019xlnet}, and RoBERTa \cite{liu2019roberta} lead tasks such as text classification, sentiment analysis, semantic role labeling, question answering, among others. However, most of the models have hundreds of millions of parameters, which significantly slows down the training process and inference time. Besides, the large number of parameters demands a lot of memory consumption, making such models hard to adopt in production environments where computational resources are strictly limited.

% PARAGRAPH MESSAGE: Teacher-student setting for model compression is good
Due to these limitations, many approaches have been proposed to reduce the size of the models while still providing similar performance. One of the most effective techniques is knowledge distillation (KD) in a teacher-student setting \cite{hinton2015distilling}, where a cumbersome already-optimized model (i.e., the teacher) produces output probabilities that are used to train a simplified model (i.e., the student). Unlike training with one-hot labels where the classes are mutually exclusive, using a probability distribution provides more information about the similarities of the samples, which is the key part of the teacher-student distillation.

% PARAGRAPH MESSAGE: Teacher-student KD is good but has some problems
Even though the student requires fewer parameters while still performing similar to the teacher, recent work shows the difficulty of distilling information from a huge model. \citeauthor{mirzadeh2019improved} \shortcite{mirzadeh2019improved} state that, when the gap in between the teacher and the student is large (e.g., shallow vs. deep neural networks), the student struggles to approximate the teacher. They propose to use an intermediate teaching assistant (TA) model to distill the information from the teacher and then use the TA model to distill information towards the student. However, we argue that the abstraction captured by a large teacher is only exposed through the output probabilities, which makes the internal knowledge from the teacher (or the TA model) hard to infer by the student. This can potentially take the student to very different internal representations undermining the generalization capabilities initially intended to be transferred from the teacher.

% PARAGRAPH MESSAGE: Internal knowledge distillation is even better!
In this paper, we propose to apply KD to internal representations. Our approach allows the student to internally behave as the teacher by effectively transferring its linguistic properties. We perform the distillation at different internal points across the teacher, which allows the student to learn and compress the abstraction in the hidden layers of the large model systematically. By including internal representations, we show that our student outperforms its homologous models trained on ground-truth labels, soft-labels, or both.

% ========================================================================
\section{Related Work}
% ========================================================================

% PARAGRPAH MESSAGE: Previous work extending KD as a framework on different scenarios
Knowledge distillation has become one of the most effective and simple techniques to compress huge models into simpler and faster models. The versatility of this framework has allowed the extension of KD to scenarios where a set of expert models in different tasks distill their knowledge into a unified multi-task learning network \cite{clark2019bam}, as well as the opposite scenario where an ensemble of multi-task models are distilled into a task-specific network \cite{liu2019mt-dnn-kd,liu2019mt-dnn}. We extend the knowledge distillation framework with a different formulation by applying the same principle to internal representations.

Using internal representations to guide the training of a student model was initially explored by \citeauthor{romero2014fitnets} \shortcite{romero2014fitnets}. They proposed \textsc{FitNet}, a convolutional student network that is thinner and deeper than the teacher while using significantly fewer parameters. In their work, they establish a middle point in both the teacher and the student models to compare internal representations. Since the dimensionality between the teacher and the student differs, they use a convolutional regressor model to map such vectors into the same space, which adds a significant number of parameters to learn. Additionally, they mainly focus on providing a deeper student network than the teacher, exploiting the particular benefits of depth in convolutional networks. Our work differs from theirs in different aspects: 
1) using a single point-wise loss on the middle layers has mainly a regularization effect, but it does not guarantee to transfer the internal knowledge from the teacher; 
2) our distillation method is applied across all the student layers, which effectively compress groups of layers from the teacher into a single layer of the student; 
3) we use the internal representations as-is instead of relying on additional parameters to perform the distillation; 
4) we do not focus on deeper models than the teacher as this can slow down the inference time, and it is not necessarily an advantage on transformer-based models.

% TODO: add the concurrent related work here
Concurrent to this work, similar transformer-based distillation techniques have been studied. \citeauthor{sanh2019distilbert} \shortcite{sanh2019distilbert} propose DistilBERT, which compresses BERT during pre-training to provide a smaller general-purpose model. They pre-train their model using a masked language modeling loss, a cosine embedding loss at the hidden states, and the teacher-student distillation loss. Conversely, \citeauthor{sun-etal-2019-patient} \shortcite{sun-etal-2019-patient} distill their model during task-specific fine-tuning using a MSE loss at the hidden states and cross-entropy losses from soft- and hard-labels. While our work is similar to theirs, the most relevant differences are 1) the use of KL-divergence loss at the self-attention probabilities, which have been shown to capture linguistic knowledge \cite{clark2019what}, and 2) the introduction of new algorithms to distill the internal knowledge from the teacher (i.e., progressive and stacked knowledge distillation).

Curriculum learning (CL) \cite{bengio2009learning} is another line of research that focuses on teaching complex tasks by building upon simple concepts. Although the goal is similar to ours, CL is conducted by stages focusing on simple tasks first and progressively moving to more complicated tasks. However, this method requires annotations among the preliminary tasks, and they have to be carefully picked so that the order and relation among the build-up tasks are helpful for the model. Unlike CL, we focus on teaching the internal representations of an optimized complex model, which are assumed to have the preliminary build-up knowledge for the task of interest.

% TODO: mention other model compression techniques such as pruning and quantization
Other model compression techniques include quantization \cite{hubara2017quantized,he2016effectiveQuantization,courbariaux2016binarized} and weights pruning \cite{han2015deep}. The first one focuses on approximating a large model into a smaller one by reducing the precision of each of the parameters. The second one focuses on removing weights in the network that do not have a substantial impact on model performance. These techniques are complementary to the method we propose in this paper, which can potentially lead to a more effective overall compression approach. 

% ========================================================================
\section{Methodology}
% ========================================================================

In this section, we detail the process of distilling knowledge from internal representations. First, we describe the standard KD framework \cite{hinton2015distilling}, which is an essential part of our method. Then, we formalize the objective functions to distill the internal knowledge of transformer-based models. Lastly, we propose various algorithms to conduct the internal distillation process.

% ==============================================
\subsection{Knowledge Distillation}
% ==============================================

\citeauthor{hinton2015distilling} \shortcite{hinton2015distilling} proposed knowledge distillation (KD) as a framework to compress a large model into a simplified model that achieves similar results. The framework uses a teacher-student setting where the student learns from both the ground-truth labels (if available) and the soft-labels provided by the teacher. The probability mass associated with each class in the soft-labels allows the student to learn more information about the label similarities for a given sample. The formulation of KD considering both soft and hard labels is given as follows:
\begin{align} \label{eq:kd_hinton}
	\mathcal{L}_{KD} =& - \frac{1}{N}\sum_i^N p(y_i|x_i, \theta_T) log(\hat{y}_i) - \lambda\frac{1}{N}\sum_i^N y_i log(\hat{y}_i))
\end{align}
\noindent where $\theta_T$ represents the parameters of the teacher, and $p(y_i|x_i, \theta_T)$ are its soft-labels; $\hat{y_i}$ is the student prediction given by $p(y_i|x_i, \theta_S)$ where $\theta_S$ denotes its parameters, and $\lambda$ is a small scalar that weights down the hard-label loss. Since the soft-labels often present high entropy, the gradient tends to be smaller than the one from the hard-labels. Thus, $\lambda$ balances the terms by reducing the impact of the hard loss. 

% ==============================================
\subsection{Matching Internal Representations}
% ==============================================

\begin{figure}[t!]
	\centering
	\includegraphics[width=0.78\linewidth]{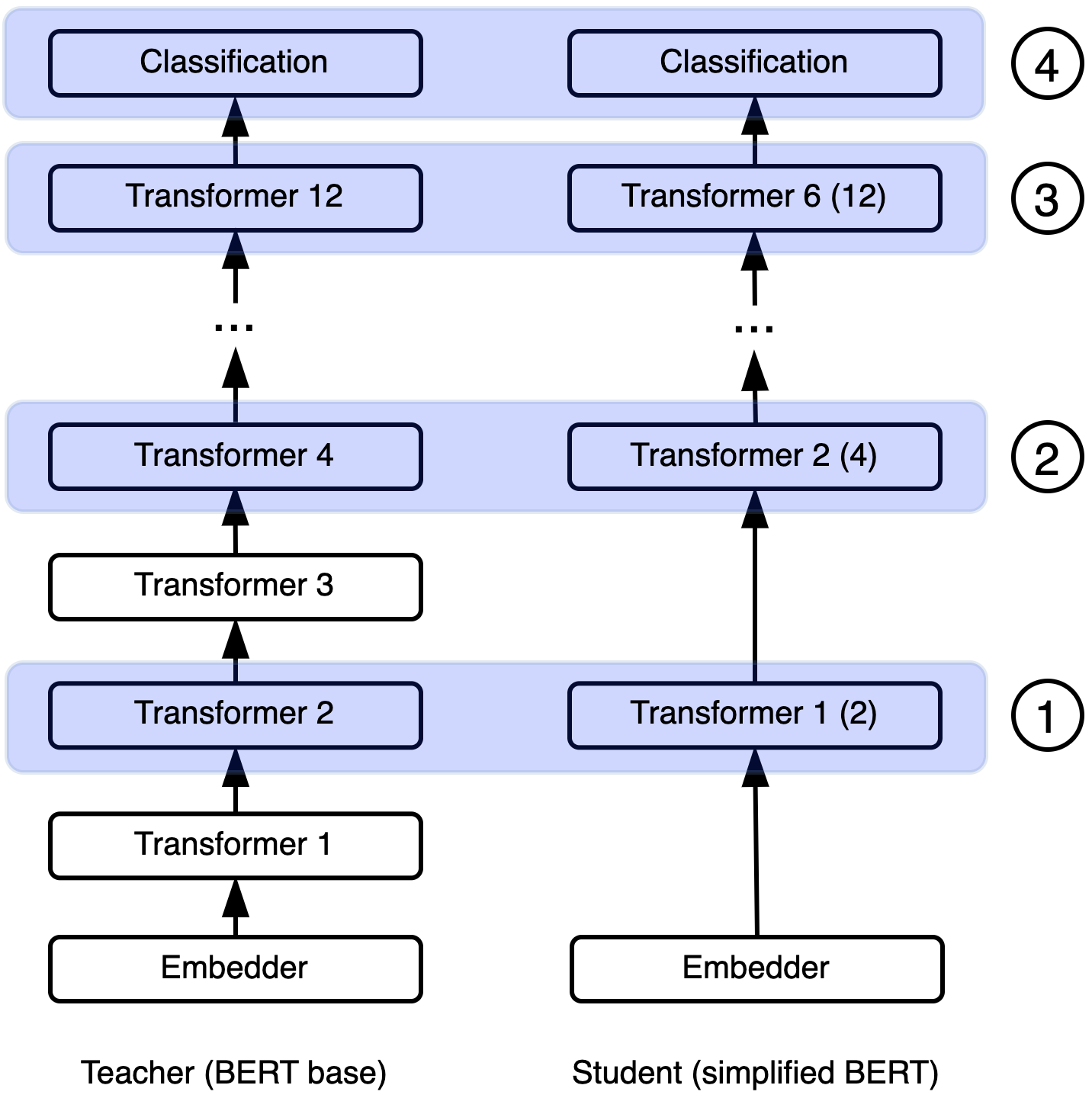}
    \caption{Knowledge distillation from internal representations. We show the internal layers that the teacher (left) distills into the student (right).}
	\label{fig:algorithm_progressive_kd}
\end{figure}

In order to make the student model behave as the teacher model, the student is optimized by the soft-labels from teacher's output. In addition, the student also acquires the abstraction hidden in the teacher by matching its internal representations. 
That is, we want to teach the student how to internally behave by compressing the knowledge of multiple layers from the teacher into a single layer of the student. Figure \ref{fig:algorithm_progressive_kd} shows a teacher with twice the number of layers of the student, where the colored boxes denote the layers where the student is taught the internal representation of the teacher. In this case, the student compresses two layers into one while preserving the linguistic behavior across the teacher layers.

We study the internal KD of transformer-based models, specifically the case of BERT and simplified versions of it (i.e., fewer transformer layers). We define the internal KD by using two terms in the loss function. Given a pair of transformer layers to match (see Figure \ref{fig:algorithm_progressive_kd}), we calculate (1) the Kullback-Leibler (KL) divergence loss across the self-attention probabilities of all the transformer heads\footnote{We are interested in a loss function that considers the probability distribution as a whole, and not point-wise errors.}, and (2) the cosine similarity loss between the \texttt{[CLS]} activation vectors for the given layers.

\textbf{KL-divergence loss}. Consider $A$ as the self-attention matrix that contains row-wise probability distributions per token in a sequence given by $A = \mathrm{softmax}(d^{-0.5}_a QK^T)$ \cite{vaswani2017attention}. For a given head in a transformer layer, we use the KL-divergence loss as follows:
\begin{align} \label{eq:kl_div_loss}
	\mathcal{L}_{kl} &= \frac{1}{L}\sum_i^L A_{T_{i}} log\frac{A_{T_{i}}}{A_{S_{i}}}
\end{align}
\noindent where $L$ is the length of a sequence, $A_{T_{i}}$ and $A_{S_{i}} \ $ describe the $i$-th row of the self-attention matrix for the teacher and student, respectively. The motivation of applying this loss function to the self-attention matrices comes from recent research that documents the linguistic patterns captured by the attention probabilities of BERT \cite{clark2019what}. Forcing the divergence between the self-attention probability distributions to be as small as possible preserves the linguistic behavior in the student.

\textbf{Cosine similarity loss}. For the second term of our internal distillation loss, we use cosine similarity\footnote{$L_2$ loss could be used as well without impacting generality.} as follows:
\begin{align} \label{eq:cos_sim_loss}
    \mathcal{L}_{cos} = 1 - cos(h_T, h_S)
\end{align}
where $h_T$ and $h_S$ are the hidden vector representations for the \texttt{[CLS]} token for the teacher and student, respectively. We include this term in our internal KD formulation to consider a similar behavior in the activation going through the network. That is, while KL-divergence focuses on the self-attention matrix, it is the weighted hidden vectors that finally pass to the upper layers, not the probabilities. Even if we force the self-attention probabilities to be similar, there is no guarantee that the final activation passed to the upper layers is similar. Thus, using this extra term, we can regularize the context representation of the sample to be similar to the one from the teacher.\footnote{We only use the context vector instead of all the hidden token vectors to avoid over-regularizing the model \cite{romero2014fitnets}.}

% Documentation: http://tug.ctan.org/macros/latex/contrib/algorithmicx/algorithmicx.pdf
\begin{algorithm}
    \caption{Stacked Internal Distillation (SID) }
    \label{alg:internal_distillation}
    \begin{algorithmic}[1]
        \small
        \Procedure{headLoss}{$\text{TS}, batch, layer_t, layer_s$}
            \For{$sample \gets batch; \text{\textbf{init}}~\mathcal{L} \gets 0;$}
                \State $\mathrm{P} \gets$ \Call{concatHeads}{$\text{TS.teacher}, sample, layer_t$}
                \State $\mathrm{Q} \gets$ \Call{concatHeads}{$\text{TS.student}, sample, layer_s$}
                \State $\mathcal{L} \gets \mathcal{L} + mean(sum(\mathrm{P} * log(\mathrm{P} / \mathrm{Q}), \text{axis=}2))$ 
            \EndFor
            \State \textbf{return} $\mathcal{L} / size(batch)$
        \EndProcedure
        % \Statex
        \Procedure{StackIntDistill}{$\text{TS}, batch$}
            \State $\mathcal{L}_{cos}, \mathcal{L}_{kl} \gets 0, 0$ 
            \For{$layer_t, layer_s \gets \Call{match}{0 \dots \text{TS.nextLocked})}$}
                \State $\mathrm{CLS}_{t} \gets$ \Call{getCLS}{$\text{TS.teacher}, batch, layer_t$}
                \State $\mathrm{CLS}_{s} \gets$ \Call{getCLS}{$\text{TS.student}, batch, layer_s$}
                \State $\mathcal{L}_{cos} \gets \mathcal{L}_{cos} + mean(1 - cos(\mathrm{CLS}_{t}, \mathrm{CLS}_{s}))$
                \State $\mathcal{L}_{kl} \gets \mathcal{L}_{kl}$ + \Call{headLoss}{$\text{TS}, batch, layer_t, layer_s$}
            \EndFor
            \State \textbf{return} $\mathcal{L}_{cos}, \mathcal{L}_{kl}$
        \EndProcedure
        % \Statex
        \State $\text{TS} \gets \Call{InitializeTSModel}$
        \Repeat{:}
            \For{$e \gets 0, epochs$}
                \If{$\text{TS.nxtLockedLay} < \text{TS.student.nLayers}$}
                    \LineComment{Perform internal distillation}
                    \For{$batch \gets data; ~\text{\textbf{init}}~\tau \gets 0;$}
                        \State $\mathcal{L}_{cos}, \mathcal{L}_{kl} \gets$ \Call{StackIntDistill}{$\text{TS}, batch$}
                        \State $backprop(\text{TS}, \mathcal{L}_{cos} + \mathcal{L}_{kl})$
                        \State $\tau \gets \tau + \mathcal{L}_{cos}$ \Comment{Accumulate for threshold}
                    \EndFor
                    \If{$\tau < \mathcal{T}$ OR $e \ge lim(\text{TS.nxtLockedLay}, e)$}
                        \State $\text{TS.nextLockedLay} \gets \text{TS.nxtLockedLay} + 1$
                    \EndIf
                \Else
                    \LineComment{Perform standard distillation}
                    \For{$batch \gets data$}
                        % \State $\mathcal{L}_{xent} \gets$ \Call{CrossEntropy}{\text{TS.forward}(batch)}
                        % \State $backprop(\text{TS}, \mathcal{L}_{xent})$
                        \State $backprop(\text{TS}, xentropy(\text{TS}, batch))$
                    \EndFor
                \EndIf
            \EndFor
        \Until{convergence}
    \end{algorithmic}
\end{algorithm}

% ==============================================
\subsection{How to Distill the Internal Knowledge?}
% ==============================================

Different layers across the teacher capture different linguistic concepts. Recent research shows that BERT builds linguistic properties that become more complex as we move from the bottom to the top of the network \cite{clark2019what}. Since the model builds upon bottom representations, in addition to distilling all the internal layers simultaneously, we also consider distilling knowledge progressively matching internal representation in a bottom-up fashion. More specifically, we consider the following scenarios:

\begin{enumerate}
    \item \textbf{Internal distillation of all layers}. All the layers of the student are optimized to match the ones from the teacher in every epoch. In Figure \ref{fig:algorithm_progressive_kd}, the distillation simultaneously occurs on the circled numbers \circled{1}, \circled{2}, \circled{3}, and \circled{4}. 
    \item \textbf{Progressive internal distillation (PID)}. We distill the knowledge from lower layers first (close to the input) and progressively move to upper layers until the model focuses only on the classification distillation. Only one layer is optimized at a time. In Figure \ref{fig:algorithm_progressive_kd}, the loss will be given by the transition \circled{1} $\rightarrow$ \circled{2} $\rightarrow$ \circled{3} $\rightarrow$ \circled{4}.
    \item \textbf{Stacked internal distillation (SID)}. We distill the knowledge from lower layers first, but instead of moving from one layer to another exclusively, we keep the loss produced by previous layers stacking them as we move to the top. Once at the top, we only perform classification (see Algorithm \ref{alg:internal_distillation}). In Figure \ref{fig:algorithm_progressive_kd}, the loss is determined by the transition \circled{1} $\rightarrow$ \circled{1} + \circled{2} $\rightarrow$ \circled{1} + \circled{2} + \circled{3} $\rightarrow$ \circled{4}.
\end{enumerate}

For the last two scenarios, to move to upper layers, the student either reaches a limited number of epochs per layer or a cosine loss threshold, whatever happens first (see line 24 in Algorithm \ref{alg:internal_distillation}). Additionally, these two scenarios can be combined with the classification loss at all times, not only until the model reaches the top layer.

\begin{table*}[t!]
\centering
% \renewcommand{\arraystretch}{0.7}
% \setlength{\tabcolsep}{8.51pt}
% \small
\begin{tabular}{llcccc}
\toprule
\multirow{2}{*}{\textbf{Experiment}} 
& \multirow{2}{*}{\textbf{Description}} 
& \textbf{CoLA {[}8.5k{]}}
& \textbf{QQP {[}364k{]}}
& \textbf{MRPC {[}3.7k{]}}
& \textbf{RTE {[}2.5k{]}}          \\
&
& \multicolumn{1}{c}{\textbf{MCC}} 
& \multicolumn{1}{c}{\textbf{Acuracy / F1}} 
& \multicolumn{1}{c}{\textbf{Acuracy / F1}} 
& \multicolumn{1}{c}{\textbf{Acuracy}} \\
\midrule
\multicolumn{6}{l}{\textit{Fine-tuning BERT\textsubscript{base} and BERT\textsubscript{6} without KD}} \\
\midrule
Exp1.0  & BERT\textsubscript{base}             & 60.16     & 91.44 / 91.45     & 83.09 / 82.96     & 67.51    \\
Exp1.1  & BERT\textsubscript{6}   & 44.56     & 90.58 / 90.62     & 76.23 / 73.72     & 59.93    \\
\midrule
\multicolumn{6}{l}{\textit{Fine-tuning BERT\textsubscript{6} with different KD techniques using BERT\textsubscript{base} (Exp1.0) as teacher}}   \\
\midrule
Exp2.0  & BERT\textsubscript{6} soft                                 & 41.72     & 90.61 / 90.65     & 77.21 / 75.74     & 62.46   \\
Exp3.0  & BERT\textsubscript{6} soft + kl                            & 43.70     & 91.32 / 91.32     & \bf 83.58 / 82.46 & 67.15   \\
Exp3.1  & BERT\textsubscript{6} soft + cos                           & 42.64     & 91.08 / 91.10     & 79.66 / 78.35     & 57.04   \\
Exp3.2  & BERT\textsubscript{6} soft + kl + cos                      & 42.07     & \bf 91.37 / 91.38 & 83.09 / 81.39     & 66.43   \\
Exp3.3  & BERT\textsubscript{6} [PID] kl + cos $\rightarrow$ soft    & 45.54     & 91.22 / 91.24     & 81.62 / 80.12     & 64.98   \\
Exp3.4  & BERT\textsubscript{6} [SID] kl + cos $\rightarrow$ soft    & \bf 46.09 & 91.25 / 91.27     & 82.35 / 81.39     & 64.62   \\
Exp3.5  & BERT\textsubscript{6} [SID] kl + cos + soft                & 43.93     & 91.21 / 91.22     & 81.37 / 79.16     & 66.43   \\
Exp3.6  & BERT\textsubscript{6} [SID] kl + cos + soft + hard         & 42.55     & 91.20 / 91.21     & 70.10 / 69.68     & \bf 67.51   \\
\bottomrule
\end{tabular}
\caption{\label{tab:dev_results}The development results across four datasets. Experiments 1.0 and 1.1 are trained without any distillation method, whereas experiments 2.0 and 3.X use a different combination of algorithms to distill information. Experiment 2.0 only uses standard knowledge distillation, and it can be considered as baseline.}
\end{table*}

% ========================================================================
\section{Experiments and Results}
% ========================================================================

\subsection{Datasets}

We conduct experiments on four datasets of the GLUE benchmark \cite{wang2018glue}, which we describe briefly: 
\begin{enumerate}
    \item \textbf{CoLA}. The Corpus of Linguistic Acceptability \cite{warstadt2018cola} is part of the single sentence tasks, and it requires to determine whether an English text is grammatically correct. It uses the Matthews Correlation Coefficient (MCC) to measure the performance.
    
    \item \textbf{QQP}. The Quora Question Pairs\footnote{\url{data.quora.com/First-Quora-Dataset-Release-Question-Pairs}} is a semantic similarity dataset, where the task is to determine whether two questions are semantically equivalent or not. It uses accuracy and F1 as metrics.
    
    \item \textbf{MRPC}. The Microsoft Research Paraphrase Corpus \cite{dolan2005mrpc} contains pairs of sentences whose annotations describe whether the sentences are semantically equivalent or not. Similar to QQP, it uses accuracy and F1 as metrics.
    
    \item \textbf{RTE}. The Recognizing Textual Entailment \cite{wang2018glue} has a collection of sentence pairs whose annotations describe entitlement or not entitlement between the sentences (formerly annotated with labels entitlement, contradiction or neutral). It uses accuracy as a metric.
    % \item \textbf{QNLI}. This dataset uses the Stanford Question Answering Dataset \cite{rajpurkar2016qnli}. The GLUE benchmark formulates the task as a sentence pair classification where the goal is to determine whether a context sentence contains the answer to a given question.
\end{enumerate}

For the MRPC and QQP datasets, the metrics are accuracy and F1, but we optimize the models on F1 only.

\subsection{Parameter Initialization}

We experiment with BERT\textsubscript{base} \cite{devlin2018bert} and simplified versions of it. In the case of BERT with 6 transformer layers, we initialize the parameters using different layers of the original BERT\textsubscript{base} model, which has 12 transformer layers. Since our goal is to compress the behavior of a subset of layers into one layer, we initialize a layer of the simplified BERT model with the upper layer of the subset. For example, Figure \ref{fig:algorithm_progressive_kd} shows the compression of groups of two layers into one layer, hence, the first layer of the student model is initialized with the parameters of the second layer of the BERT\textsubscript{base} model.\footnote{Note that the initialization does not take the parameters of the fine-tuned teacher. Instead, we use the parameters of the general-purpose BERT\textsubscript{base} model.}

\subsection{Experimental Setup}

Table \ref{tab:dev_results} shows the results on the development set across four datasets. We define the experiments as follows:
\begin{itemize}
    \item \textbf{Exp1.0: BERT\textsubscript{base}}. This is the standard BERT\textsubscript{base} model that is fine-tuned on task-specific data without any KD technique. Once optimized, we use this model as a teacher for the KD experiments.
    
    \item \textbf{Exp1.1: BERT\textsubscript{6}}. This is a simplified version of BERT\textsubscript{base}, where we use 6 transformer layers instead of 12. The layer selection for initialization is described in the previous section. We do not use any KD for this experiment. The KD experiments described below use this architecture as the student model.
    
    \item \textbf{Exp2.0: BERT\textsubscript{6} soft}. The model is trained with soft-labels produced by the fine-tuned BERT\textsubscript{base} teacher from experiment 1.0. This scenario corresponds to Equation \ref{eq:kd_hinton} with $\lambda=0$ to ignore the one-hot loss.

    \item \textbf{Exp3.0: BERT\textsubscript{6} soft + kl}. The model uses both the soft-label and the KL-divergence losses from Equations \ref{eq:kd_hinton} and \ref{eq:kl_div_loss}. The KL-divergence loss is averaged across all the self-attention matrices from the student (i.e., 12 attention heads per transformer layer per 12 transformer layers).
    
    \item \textbf{Exp3.1: BERT\textsubscript{6} soft + cos}. The model uses both the soft-label and the cosine similarity losses from Equations \ref{eq:kd_hinton} and \ref{eq:cos_sim_loss}. The cosine similarity loss is computed from the \texttt{[CLS]} vector from all matching layers.
    
    \item \textbf{Exp3.2: BERT\textsubscript{6} soft + kl + cos}. The model uses all the losses from all layers every epoch. This experiment combines experiments 3.0 and 3.1.
    
    \item \textbf{Exp3.3: BERT\textsubscript{6} [PID] kl + cos $\rightarrow$ soft}. The model only uses \textit{progressive internal distillation} until it reaches the classification layer. Once there, only soft-labels are used.
    
    \item \textbf{Exp3.4: BERT\textsubscript{6} [SID] kl + cos $\rightarrow$ soft}. The model uses \textit{stacked internal distillation} until it reaches the classification layer. Once there, only soft-labels are used.
    
    \item \textbf{Exp3.5: BERT\textsubscript{6} [SID] kl + cos + soft}. The model uses \textit{stacked internal distillation} and soft-labels distillation all the time during training.
    
    \item \textbf{Exp3.6: BERT\textsubscript{6} [SID] kl + cos + soft + hard}. Same as Exp3.5, but it includes the hard-labels in the Equation \ref{eq:kd_hinton} with $\lambda = 0.1$.
\end{itemize}

We optimize our models using Adam with an initial learning rate of 2e-5 and a learning rate scheduler as described by \citeauthor{devlin2018bert} \shortcite{devlin2018bert}. We fine-tune BERT\textsubscript{base} for 10 epochs, and the simplified BERT models for 50 epochs both with a batch size of 32 samples and a maximum sequence length of 64 tokens. We evaluate the statistical significant of our models using t-tests as described by  \citeauthor{rotem2018significance} \shortcite{rotem2018significance}. All the internal KD results have shown statistical significance with a p-value less than 1e-3 with respect to the standard KD method across the datasets.

\subsection{Development and Evaluation Results}

\begin{table}[t!]
\centering
\small
\begin{tabular}{lcccc}
\toprule
\multirow{2}{*}{\textbf{Exp.}} 
& \textbf{CoLA} & \textbf{QQP} & \textbf{MRPC} & \textbf{RTE} \\
& \textbf{MCC} & \multicolumn{1}{c}{\textbf{Acc. / F1}} & \multicolumn{1}{c}{\textbf{Acc. / F1}} & \multicolumn{1}{c}{\textbf{Acc.}} \\
\midrule
Exp1.0  & 51.4  & 71.3 / 89.2   & 84.9 / 79.9   & 66.4   \\
Exp2.0  & 38.3  & 69.1 / 88.0   & 81.6 / 73.9   & 59.7   \\
Exp3.X  & 41.4  & 70.9 / 89.1   & 83.8 / 77.1   & 62.2  \\
\bottomrule
\end{tabular}
\caption{\label{tab:test_results} The test results from the best models according to the development set. We add Exp1.0 (BERT\textsubscript{base}) for reference. Exp2.0 uses BERT\textsubscript{6} with standard distillation (soft-labels only), and Exp3.X uses the best internal KD technique with BERT\textsubscript{6} as student according to the development set. 
% Note that BERT\textsubscript{6} has 66.9M parameters while BERT base has 109.4M.
}
\end{table}

As shown in Table \ref{tab:dev_results}, we perform extensive experiments for BERT\textsubscript{6} as a student, where we evaluate different training techniques with or without knowledge distillation. In general, the first thing to notice is that the distillation techniques outperforms BERT\textsubscript{6} trained without distillation (Exp1.1). While it is not always the case for standard distillation (Exp1.1 vs. Exp2.0 for CoLA), the internal distillation method proposed in this work consistently outperforms both Exp1.1 and Exp2.0 across all datasets. Nevertheless, the gap between the results substantially depends on the size of the data. Intuitively, this is expected behavior since the more data we provide to the teacher, the more knowledge is exposed, and hence, the student reaches a more accurate approximation of the teacher.

Additionally, our internal distillation results are consistently better than the standard soft-label distillation in the test set, as described in Table \ref{tab:test_results}.

% ========================================================================
\section{Analysis}
% ========================================================================

This section provides more insights into our algorithm based on parameter reduction, data size impact, model convergence, self-attention behavior, and error analysis.

% ==============================================
\subsection{Performance vs. Parameters}
% ==============================================

\begin{figure}[t!]
	\centering
	\includegraphics[width=0.9\linewidth]{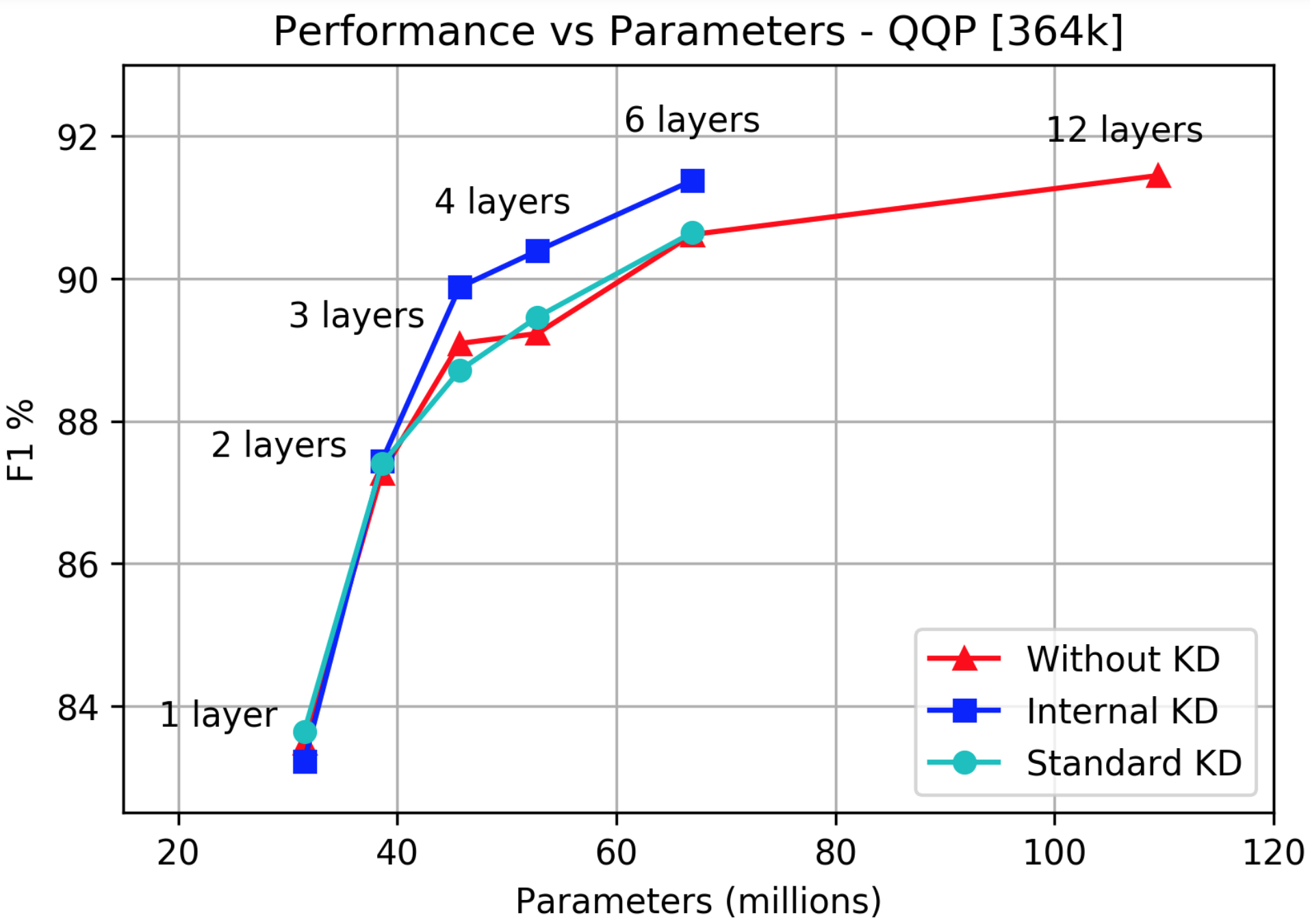}
    \caption{ Performance vs. parameters trade-off. The points along the lines denote the number of layers used in BERT, which is reflected by the number of parameters in the x-axis. }
	\label{fig:performance_all}
\end{figure}

% TODO: also compare with standard distillation since I have the numbers too!
We analyze the parameter reduction capabilities of our method. Figure \ref{fig:performance_all} shows that BERT\textsubscript{6} can easily achieve similar results than the original BERT\textsubscript{base} model with 12 transformer layers. Note that BERT\textsubscript{base} has around 109.4M parameters, which can be broken down into 23.8M parameters related to embeddings and around 85.6M parameters related to transformer layers. The BERT\textsubscript{6} student, however, has 43.1M parameters in the transformer layers, which means that the parameter reduction is about 50\%, while still performing very similar to the teacher (91.38 F1 vs. 91.45 F1 for QQP, see Table \ref{tab:dev_results}). Also, note that the 0.73\% F1 drop is statistical significant between the student only trained on soft-labels and the student trained with our method. 
% This behavior is also seen across multiple datasets (see Appendix).
% the drop in performance is consistent across datasets.

% Embedding parameters  =  23,837,184  (23.8M)
% BERT_base parameters  = 109,483,009 (109.4M)
% BERT_6                =  66,955,777  (66.9M)
% 
% BERT_base - Embedding = 85,645,825  (85.6M)
% BERT_6    - Embedding = 43,118,593  (43.1M)
% 43,118,593 / 85,645,825 = 0.5034  ---> 50% Reduction in the actual parameter target

Moreover, if we keep reducing the number of layers, the performance decays for both student models (see Figure \ref{fig:performance_all}). However, the internal distillation method is more resilient to keep a higher performance. Eventually, with one transformer layer to distill internally, the compression rate is too high for the model to account for an additional boost when we compare BERT\textsubscript{1} students with standard and internal distillation methods.

% ==============================================
\subsection{The Impact of Data Size}
% ==============================================

\begin{figure}[t!]
	\centering
	\includegraphics[width=0.9\linewidth]{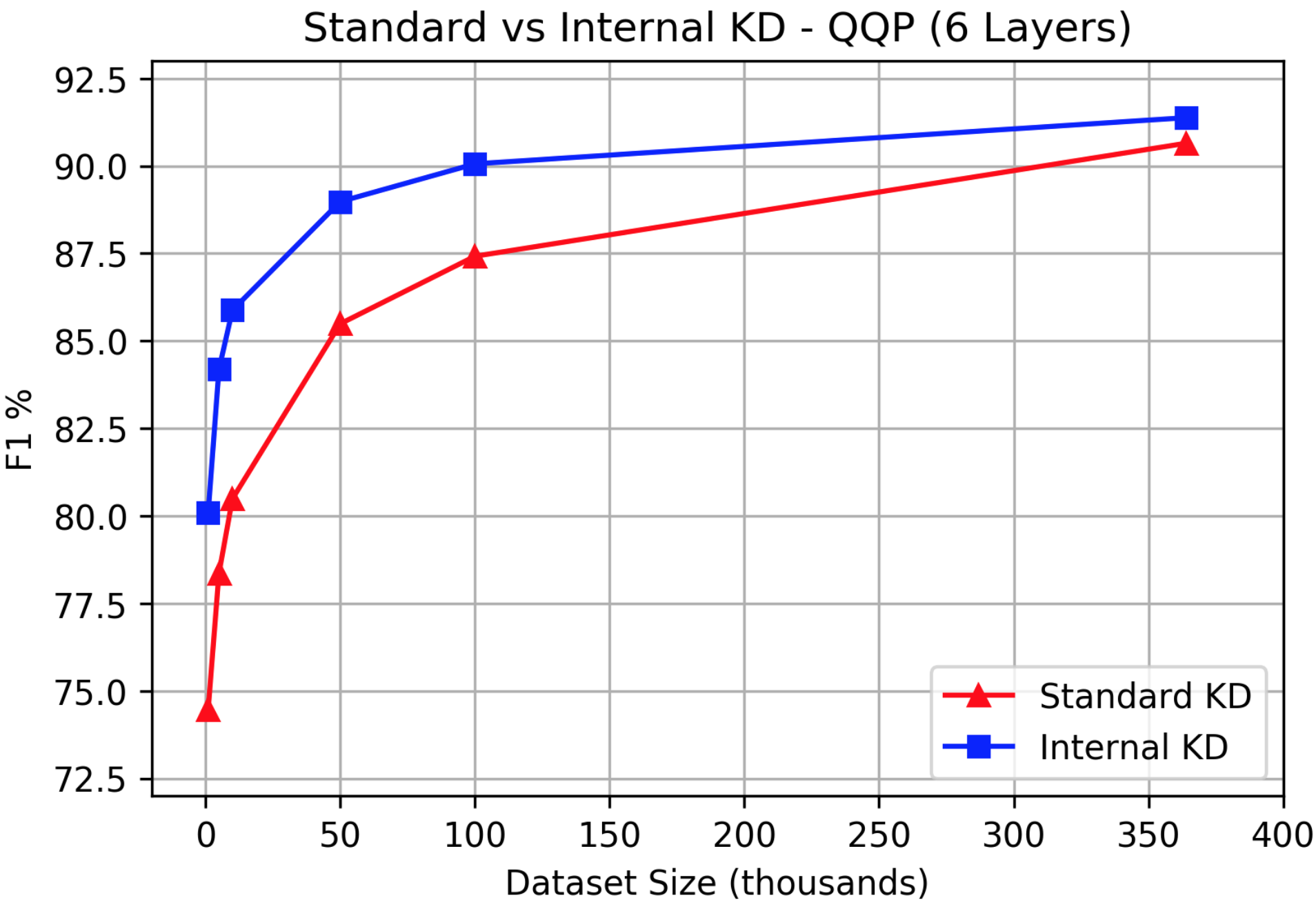}
	\caption{ The impact of training size for standard vs. internal KD. We experiment with sizes between 1K and +350K. }
	\label{fig:datasize_all}
\end{figure}

We also evaluate the impact of the data size. For this analysis, we fix the student architecture to the BERT\textsubscript{6}, and we only modify the size of the training data. We compare the standard and the internal distillation techniques for the QQP dataset, as shown in Figure \ref{fig:datasize_all}. Consistently, the internal distillation outperforms the soft-label KD method. However, the gap between the two methods is small when the data size is large, but it tends to increase in favor of the internal KD method when the data size decreases.

% ==============================================
\subsection{Student Convergence}
% ==============================================

\begin{figure}[t!]
    \centering
    \includegraphics[width=0.9\linewidth]{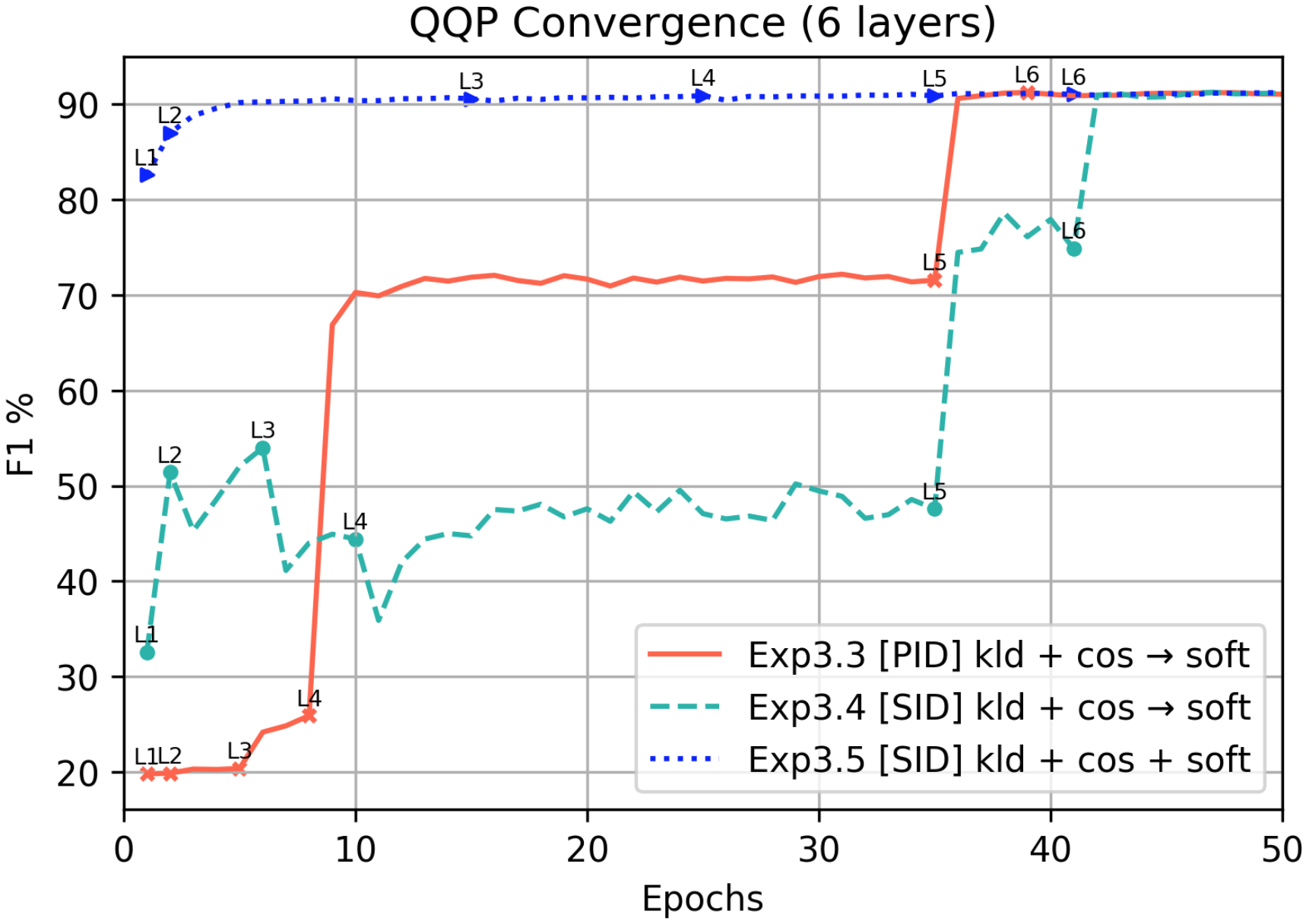}
    \caption{ Comparing algorithm convergences across epochs. The annotations along the lines denote the layers that have been completely optimized. After the L6 point, only the classification layer is trained.}
    \label{fig:convergence_qqp_all}
\end{figure}

We analyze the convergence behavior during training by comparing the performance of the internal distillation algorithms across epochs. We conduct the experiments on the QQP dataset as described in Figure \ref{fig:convergence_qqp_all}. We control over the student architecture, which is BERT\textsubscript{6}, and exclusively experiment with different internal KD algorithms. 
The figure shows three experiments: \textit{progressive internal distillation} (Exp3.3), \textit{stacked internal distillation} (Exp3.4), and \textit{stacked internal distillation} using soft-labels all the time (Exp3.5).
Importantly, note that Exp3.3 and Exp3.4 do not update the classification layer until around epoch 40 when all the transformer layers have been optimized. 
Nevertheless, the internal distillation by itself allows the students to reach higher performance across epochs eventually. In fact, Exp3.3 reaches its highest value when the 6th transformer layer is being optimized while the classification layer remains as it was initialized (see epoch 38 in Figure \ref{fig:convergence_qqp_all}).
This serves as strong evidence that the internal knowledge of the model can be taught and compress without even considering the classification layer.

% ==============================================
\subsection{Inspecting the Attention Behavior}
% ==============================================

\begin{figure*}[t!]
    \centering
    \includegraphics[width=0.89\linewidth]{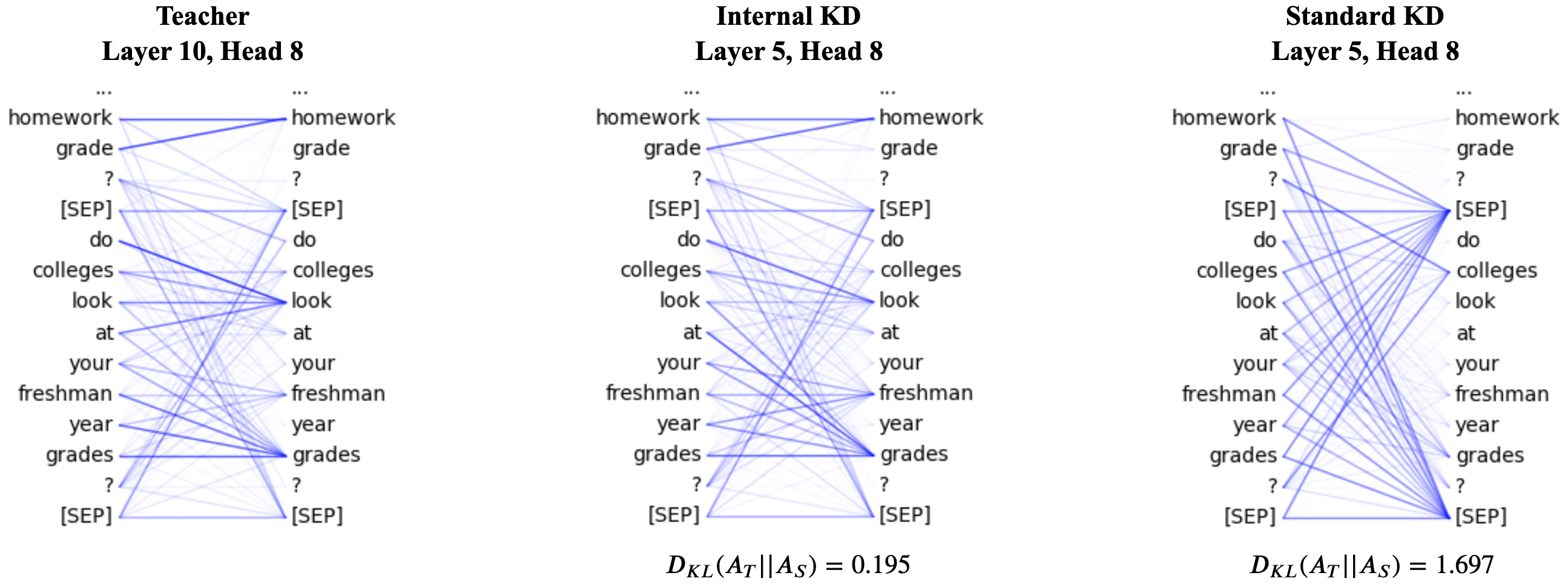}
    \caption{Attention comparison for head 8 in layer 5, each student with its corresponding head KL-divergence loss. The KL-divergence loss for the given example across all matching layers between the students and the teacher is 2.229 and 0.085 for the standard KD and internal KD students, respectively.}
    \label{fig:attention_qqp}
\end{figure*}

We inspect the internal representations learned by the students from standard and internal KD and compare their behaviors against the ones from the teacher. The goal of this experiment is to get a sense of how much the student can compress from the teacher, and how different such representations are from a student trained on soft-label in a standard KD setting. For this experiment, we use the QQP dataset and BERT\textsubscript{6} as a student. The internally-distilled student corresponds to experiment 3.2, and the soft-label student comes from experiment 2.0 (see Table \ref{tab:dev_results}). Figure \ref{fig:attention_qqp} shows the compression effectiveness of the internally distilled student with respect to the teacher. Even though the model is skipping one layer for every two layers of the teacher, the student is still able to replicate the behavior taught from the teacher. While the internal representations from the student with standard KD mainly serve to a general-purpose (i.e., attending to the separation token while spotting connections with the word college), the representations are not the ones intended to be transferred from the teacher. This means that the original goal of compressing a model does not hold entirely since its internal behavior is quite different than the one from the teacher (see Figure \ref{fig:attention_qqp} for the KL divergence on each student).

\begin{table}[t!]
\centering
\setlength{\tabcolsep}{4pt}
\small
\begin{tabular}{l|ll|ll}
\toprule
\multirow{2}{*}{\textbf{Method}} 
& \multicolumn{2}{c}{\textbf{Teacher Right}}    & \multicolumn{2}{|c}{\textbf{Teacher Wrong}} \\
& \multicolumn{2}{c}{\textbf{(36,967)}}         & \multicolumn{2}{|c}{\textbf{(3,463)}} \\
\midrule
Standard KD (Exp2.0) & 35,401 \cmark & 1,566 \xmark  & 1,232 \cmark & 2,231 \xmark \\
Internal KD (Exp3.2) & 36,191 \cmark & 776 \xmark    & 750 \cmark & 2,713 \xmark \\
\bottomrule
\end{tabular}
\caption{\label{tab:error_analysis}Right and wrong predictions on the QQP development dataset. Based on the teacher results, we show the number of right (\cmark) and wrong (\xmark) predictions by the students from standard KD (Exp2.0) and internal KD (Exp3.2).}
\end{table}
% Teacher right: 36967
%     Internal right: 36191
%     Internal wrong: 776
%     Standard right: 35401
%     Standard wrong: 1566
%     Internal right and standard wrong: 1236
%     Standard right and internal wrong: 446
% 
% Teacher wrong: 3463
%     Internal right: 750
%     Internal wrong: 2713
%     Standard right: 1232
%     Standard wrong: 2231
%     Internal right and standard wrong: 255
%     Standard right and internal wrong: 737

\begin{table*}[t!]
\centering
\small
\setlength{\tabcolsep}{3pt}
\begin{tabular}{cp{11.2cm}cccc}
\toprule
\textbf{No.} & \textbf{QQP Development Samples} & \textbf{Class} & \textbf{Teacher} & \textbf{Std KD} & \textbf{Int KD}  \\
\midrule
\multirow{2}{*}{1}  
& Q1: if donald trump loses the general election, will he attempt to seize power by force claiming the election was fraudulent? & \multirow{2}{*}{1}
& \multirow{2}{*}{1 (0.9999)} 
& \multirow{2}{*}{1 (0.9999)} 
& \multirow{2}{*}{0 (0.4221)}  \\
& Q2: how will donald trump react if and when he loses the election? & & & & \\
\midrule
\multirow{2}{*}{2}  
& Q1: can depression lead to insanity? & \multirow{2}{*}{0}
& \multirow{2}{*}{0 (0.0429)} 
& \multirow{2}{*}{0 (1.2e-4)} 
& \multirow{2}{*}{1 (0.9987)}  \\
& Q2: does stress or depression lead to mental illness? & & & & \\
\midrule
\multirow{2}{*}{3}  
& Q1: can i make money by uploading videos on youtube (if i have subscribers)? & \multirow{2}{*}{1}
& \multirow{2}{*}{1 (0.9998)} 
& \multirow{2}{*}{0 (0.0017)} 
& \multirow{2}{*}{1 (0.8868)} \\
& Q2: how do youtube channels make money? & & & & \\
\midrule
\multirow{2}{*}{4}  
& Q1: what are narendra modi's educational qualifications? & \multirow{2}{*}{0}
& \multirow{2}{*}{0 (0.0203)} 
& \multirow{2}{*}{1 (0.9999)} 
& \multirow{2}{*}{0 (0.2158)} \\
& Q2: why is pmo hiding narendra modi's educational qualifications? & & & & \\
\bottomrule
\end{tabular}
\caption{\label{tab:error_analysis_samples}Samples where the teacher predictions are right and only one of the students is wrong. We show the predicted label along with the probability for such prediction in parenthesis. We also provide the ground-truth label in the class column.}
\end{table*}

% ==============================================
\subsection{Error Analysis}
% ==============================================

In our internal KD method, the generalization capabilities of the teacher are replicated in the student model. This also implies that the student will potentially make the mistakes of the teacher. In fact, when we compare a student only trained on soft-labels (Exp2.0) against a student trained with our method (Exp3.2), we can see in Table \ref{tab:error_analysis} that the numbers of the latter align better with the teacher numbers for both wrong and right predictions. For instance, when the teacher is right (36,967), our method is right 97.9\% of the same samples (36,191), whereas the standard distillation provides a rate of 95.7\% (35,401) with more than twice the number of mistakes than our method (1,566 vs. 776). On the other hand, when the teacher is wrong (3,463), the student in our method makes more mistakes and provides less correct predictions than the student from standard KD. Nevertheless, the overall score of the student in our method significantly exceeds the score from the student trained in a standard KD setting.

We also inspect the samples where the teacher and only one of the students are right. The QQP samples 1 and 2 in Table \ref{tab:error_analysis_samples} show wrong predictions by the internally-distilled student (Exp3.2) that are not consistent with the teacher. For sample 1, although the prediction is 0, the probability output (0.4221) is very close to the threshold (0.5). Our intuition is that the internal distillation method had a regularization effect on the student such that, considering that question 2 is much more specific than question 1, it does not allow the student to tell whether is similar or not confidently. Also, it is worth noting that standard KD student is extremely confident about the prediction (0.9999), which may not be ideal since this can be a sign of over-fitting or memorization. For sample 2, although the internally-distilled student is wrong (according to ground-truth annotation and the teacher), the questions are actually related which suggests that the student model is capable of disagreeing with the teacher while still generalizing well. Samples 3 and 4 show successful cases for the internally-distilled student, while the standard KD student fails.

% ========================================================================
\section{Conclusions}
% ========================================================================

We propose a new extension of the KD method that effectively compresses a large model into a smaller one, while still preserving a similar performance from the original model. Unlike the standard KD method, where a student only learns from the output probabilities of the teacher, we teach our smaller models by also revealing the internal representations of the teacher. Besides preserving a similar performance, our method effectively compresses the internal behavior of the teacher into the student. This is not guaranteed in the standard KD method, which can potentially affect the generalization capabilities initially intended to be transferred from the teacher. Finally, we validate the effectiveness of our method by consistently outperforming the standard KD technique in four datasets of the GLUE benchmark.

\bibliographystyle{aaai}
\bibliography{aaai}

\begin{thebibliography}{}

\bibitem[\protect\citeauthoryear{Bengio}{2009}]{bengio2009learning}
Bengio, Y.
\newblock 2009.
\newblock {Learning Deep Architectures for AI}.
\newblock {\em Foundations and Trends{\textregistered} in Machine Learning}
  2(1):1--127.

\bibitem[\protect\citeauthoryear{Clark \bgroup et al\mbox.\egroup
  }{2019a}]{clark2019what}
Clark, K.; Khandelwal, U.; Levy, O.; and Manning, C.~D.
\newblock 2019a.
\newblock {What Does BERT Look At? An Analysis of BERT's Attention}.
\newblock {\em CoRR} abs/1906.04341.

\bibitem[\protect\citeauthoryear{Clark \bgroup et al\mbox.\egroup
  }{2019b}]{clark2019bam}
Clark, K.; Luong, M.; Khandelwal, U.; Manning, C.~D.; and Le, Q.~V.
\newblock 2019b.
\newblock {BAM! Born-Again Multi-Task Networks for Natural Language
  Understanding}.
\newblock {\em CoRR} abs/1907.04829.

\bibitem[\protect\citeauthoryear{Courbariaux \bgroup et al\mbox.\egroup
  }{2016}]{courbariaux2016binarized}
Courbariaux, M.; Hubara, I.; Soudry, D.; El-Yaniv, R.; and Bengio, Y.
\newblock 2016.
\newblock {Binarized Neural Networks: Training Neural Networks with Weights and
  Activations Constrained to +1 or −1}.
\newblock {\em arXiv preprint arXiv:1602.02830}.

\bibitem[\protect\citeauthoryear{Devlin \bgroup et al\mbox.\egroup
  }{2018}]{devlin2018bert}
Devlin, J.; Chang, M.-W.; Lee, K.; and Toutanova, K.
\newblock 2018.
\newblock {BERT: Pre-training of Deep Bidirectional Transformers for Language
  Understanding}.
\newblock {\em arXiv preprint arXiv:1810.04805}.

\bibitem[\protect\citeauthoryear{Dolan and Brockett}{2005}]{dolan2005mrpc}
Dolan, W.~B., and Brockett, C.
\newblock 2005.
\newblock {Automatically Constructing a Corpus of Sentential Paraphrases}.
\newblock In {\em Proceedings of the Third International Workshop on
  Paraphrasing ({IWP}2005)}.

\bibitem[\protect\citeauthoryear{Dror \bgroup et al\mbox.\egroup
  }{2018}]{rotem2018significance}
Dror, R.; Baumer, G.; Shlomov, S.; and Reichart, R.
\newblock 2018.
\newblock The hitchhiker's guide to testing statistical significance in natural
  language processing.
\newblock In {\em Proceedings of the 56th Annual Meeting of the Association for
  Computational Linguistics (Volume 1: Long Papers)},  1383--1392.
\newblock Association for Computational Linguistics.

\bibitem[\protect\citeauthoryear{Han, Mao, and Dally}{2015}]{han2015deep}
Han, S.; Mao, H.; and Dally, W.~J.
\newblock 2015.
\newblock {Deep Compression: Compressing Deep Neural Networks with Pruning,
  Trained Quantization and Huffman Coding}.
\newblock {\em arXiv preprint arXiv:1510.00149}.

\bibitem[\protect\citeauthoryear{He \bgroup et al\mbox.\egroup
  }{2016}]{he2016effectiveQuantization}
He, Q.; Wen, H.; Zhou, S.; Wu, Y.; Yao, C.; Zhou, X.; and Zou, Y.
\newblock 2016.
\newblock {Effective Quantization Methods for Recurrent Neural Networks}.
\newblock {\em CoRR} abs/1611.10176.

\bibitem[\protect\citeauthoryear{Hinton, Vinyals, and
  Dean}{2015}]{hinton2015distilling}
Hinton, G.; Vinyals, O.; and Dean, J.
\newblock 2015.
\newblock {Distilling the Knowledge in a Neural Network}.
\newblock {\em arXiv preprint arXiv:1503.02531}.

\bibitem[\protect\citeauthoryear{Hubara \bgroup et al\mbox.\egroup
  }{2017}]{hubara2017quantized}
Hubara, I.; Courbariaux, M.; Soudry, D.; El-Yaniv, R.; and Bengio, Y.
\newblock 2017.
\newblock {Quantized Neural Networks: Training Neural Networks with Low
  Precision Weights and Activations}.
\newblock {\em The Journal of Machine Learning Research} 18(1):6869--6898.

\bibitem[\protect\citeauthoryear{Lample and Conneau}{2019}]{lample2019xlm}
Lample, G., and Conneau, A.
\newblock 2019.
\newblock {Cross-lingual Language Model Pretraining}.
\newblock {\em CoRR} abs/1901.07291.

\bibitem[\protect\citeauthoryear{Liu \bgroup et al\mbox.\egroup
  }{2019a}]{liu2019mt-dnn-kd}
Liu, X.; He, P.; Chen, W.; and Gao, J.
\newblock 2019a.
\newblock {Improving Multi-Task Deep Neural Networks via Knowledge Distillation
  for Natural Language Understanding}.
\newblock {\em arXiv preprint arXiv:1904.09482}.

\bibitem[\protect\citeauthoryear{Liu \bgroup et al\mbox.\egroup
  }{2019b}]{liu2019mt-dnn}
Liu, X.; He, P.; Chen, W.; and Gao, J.
\newblock 2019b.
\newblock {Multi-Task Deep Neural Networks for Natural Language Understanding}.
\newblock In {\em Proceedings of the 57th Annual Meeting of the Association for
  Computational Linguistics},  4487--4496.
\newblock Florence, Italy: Association for Computational Linguistics.

\bibitem[\protect\citeauthoryear{Liu \bgroup et al\mbox.\egroup
  }{2019c}]{liu2019roberta}
Liu, Y.; Ott, M.; Goyal, N.; Du, J.; Joshi, M.; Chen, D.; Levy, O.; Lewis, M.;
  Zettlemoyer, L.; and Stoyanov, V.
\newblock 2019c.
\newblock {RoBERTa: {A} Robustly Optimized {BERT} Pretraining Approach}.
\newblock {\em CoRR} abs/1907.11692.

\bibitem[\protect\citeauthoryear{Mirzadeh \bgroup et al\mbox.\egroup
  }{2019}]{mirzadeh2019improved}
Mirzadeh, S.-I.; Farajtabar, M.; Li, A.; and Ghasemzadeh, H.
\newblock 2019.
\newblock {Improved Knowledge Distillation via Teacher Assistant: Bridging the
  Gap Between Student and Teacher}.
\newblock {\em arXiv preprint arXiv:1902.03393}.

\bibitem[\protect\citeauthoryear{Radford \bgroup et al\mbox.\egroup
  }{2018}]{radford2018improving}
Radford, A.; Narasimhan, K.; Salimans, T.; and Sutskever, I.
\newblock 2018.
\newblock {Improving Language Understanding by Generative Pre-Training}.
\newblock {\em URL https://s3-us-west-2. amazonaws.
  com/openai-assets/research-covers/languageunsupervised/language understanding
  paper. pdf}.

\bibitem[\protect\citeauthoryear{Radford \bgroup et al\mbox.\egroup
  }{2019}]{radford2019language}
Radford, A.; Wu, J.; Child, R.; Luan, D.; Amodei, D.; and Sutskever, I.
\newblock 2019.
\newblock {Language Models are Unsupervised Multitask Learners}.
\newblock {\em OpenAI Blog} 1(8).

\bibitem[\protect\citeauthoryear{Romero \bgroup et al\mbox.\egroup
  }{2014}]{romero2014fitnets}
Romero, A.; Ballas, N.; Kahou, S.~E.; Chassang, A.; Gatta, C.; and Bengio, Y.
\newblock 2014.
\newblock {FitNets: Hints for Thin Deep Nets}.
\newblock {\em arXiv preprint arXiv:1412.6550}.

\bibitem[\protect\citeauthoryear{Sanh \bgroup et al\mbox.\egroup
  }{2019}]{sanh2019distilbert}
Sanh, V.; Debut, L.; Chaumond, J.; and Wolf, T.
\newblock 2019.
\newblock Distilbert, a distilled version of bert: smaller, faster, cheaper and
  lighter.
\newblock {\em arXiv preprint arXiv:1910.01108}.

\bibitem[\protect\citeauthoryear{Sun \bgroup et al\mbox.\egroup
  }{2019}]{sun-etal-2019-patient}
Sun, S.; Cheng, Y.; Gan, Z.; and Liu, J.
\newblock 2019.
\newblock Patient knowledge distillation for {BERT} model compression.
\newblock In {\em Proceedings of the 2019 Conference on Empirical Methods in
  Natural Language Processing and the 9th International Joint Conference on
  Natural Language Processing (EMNLP-IJCNLP)},  4314--4323.
\newblock Hong Kong, China: Association for Computational Linguistics.

\bibitem[\protect\citeauthoryear{Vaswani \bgroup et al\mbox.\egroup
  }{2017}]{vaswani2017attention}
Vaswani, A.; Shazeer, N.; Parmar, N.; Uszkoreit, J.; Jones, L.; Gomez, A.~N.;
  Kaiser, L.; and Polosukhin, I.
\newblock 2017.
\newblock {Attention Is All You Need}.
\newblock {\em CoRR} abs/1706.03762.

\bibitem[\protect\citeauthoryear{Wang \bgroup et al\mbox.\egroup
  }{2018}]{wang2018glue}
Wang, A.; Singh, A.; Michael, J.; Hill, F.; Levy, O.; and Bowman, S.~R.
\newblock 2018.
\newblock {{GLUE:} {A} Multi-Task Benchmark and Analysis Platform for Natural
  Language Understanding}.
\newblock {\em CoRR} abs/1804.07461.

\bibitem[\protect\citeauthoryear{Warstadt, Singh, and
  Bowman}{2018}]{warstadt2018cola}
Warstadt, A.; Singh, A.; and Bowman, S.~R.
\newblock 2018.
\newblock {Neural Network Acceptability Judgments}.
\newblock {\em CoRR} abs/1805.12471.

\bibitem[\protect\citeauthoryear{Yang \bgroup et al\mbox.\egroup
  }{2019}]{yang2019xlnet}
Yang, Z.; Dai, Z.; Yang, Y.; Carbonell, J.; Salakhutdinov, R.; and Le, Q.~V.
\newblock 2019.
\newblock {XLNet: Generalized Autoregressive Pretraining for Language
  Understanding}.
\newblock {\em arXiv preprint arXiv:1906.08237}.

\end{thebibliography}

\end{document}